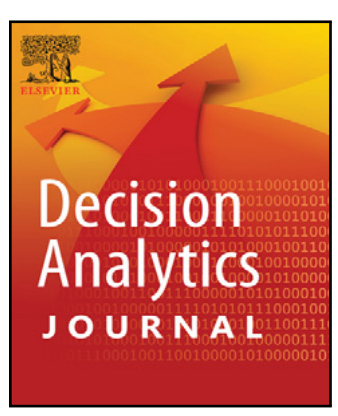

# An analysis of machine learning approaches for enhancing decision-making in complex discrete choice tasks☆

Sheng Lun (Christine) Cao [a,*], Destenie Nock [b,c], Alex Davis [d]

[a] Department of Electrical and Software Engineering, University of Calgary, Calgary, AB, Canada
[b] Department of Engineering and Public Policy, Carnegie Mellon University, Pittsburgh, PA, USA
[c] Department of Civil Engineering, Carnegie Mellon University, Pittsburgh, PA, USA
[d] Independent Decision Science Consultant, Pittsburgh, PA, USA

ABSTRACT

Discrete choice modeling is a common tool used for preference elicitation during policy-making, but this is typically done through parametric models. Machine learning can push the boundaries of discrete choice modeling for policy-based preference elicitation by adopting a data-driven approach for learning individual preferences. However, there is limited knowledge of how well machine learning methods can estimate individual discrete choice rules under individual heterogeneity, especially in the context of challenges often experienced during preference elicitation. This study evaluates four machine learning models (multinomial logistic regression, generalized additive model, twinned neural network, and Gaussian process) with respect to their capacity to learn and predict five choice rules that are important in the behavioral and social sciences (linear strong utility, monotonic strong utility, ideal point, lexicographic semiorder, and multiattribute linear ballistic accumulator). Monte Carlo experiments were performed to assess model performance when increasing a) the number of attributes in the choice alternatives, b) the number of training choice sets, and c) the choice rule's determinism. The simulation results demonstrated that semi-parametric and non-parametric models generally outperform parametric models across all choice rules and experimental contexts. Model performance also generally improves by 6% to 96% and 0% to 55%, respectively, with an increase in training choice sets and choice rule determinism. A case study using real energy policy preference data was also conducted, where TNN performed best with a BIC of 13.351. This work demonstrated the viability and limitations of semi-parametric and non-parametric models in the context of policy-centric discrete choice modeling and showed how the choice task context should drive model selection.

## 1. Introduction

Understanding how individuals make decisions between discrete options is a critical area of research, with widespread applications in, for example, transportation [1–4] and energy policy [5–7]. These decision tasks involve many complex alternatives and multiple competing objectives (e.g., minimizing costs while maximizing reliability), which present modeling difficulties. Traditionally, discrete choice modeling requires assumptions for a specific decision-making process, such as the assumption that individuals make decisions by maximizing their utility function (i.e., through a mixed logit model [8,9]), before analyzing the gathered data.

Previous studies have shown that while these parametric models are effective at recovering preferences [10,11], they are inherently limited by having to make a parametric functional form assumption about human choice behavior [12]. This study addresses the limitations behind imposing parametric modeling assumptions by investigating semi- and non-parametric machine learning approaches to infer an individual's discrete choice-making behaviors directly, allowing for greater flexibility in modeling how people make discrete choices. The goal of this work is to understand the extent to which discrete choice rules can be recovered using machine learning models, by asking: in a domain with heterogeneous preferences, how well can different machine learning models recover an individual's choice rules from a set of choices?

In order to learn an individual's choice preferences, this study integrated discrete choice rules found in decision science literature

☆ Code, additional online materials, and supplementary information (SI) are available at the project's Open Science Framework page (osf.io/cdx7e/). We have no conflicts of interest to disclose. We acknowledge the funding from the National Science Foundation, grant #2049333 (PI: Destenie Nock).

* Correspondence to: Department of Electrical and Software Engineering, University of Calgary, 2500 University Drive NW, Calgary, AB, Canada, T2N 1N4.
*E-mail address:* slcao@ucalgary.ca (S.L. Cao).

(i.e., strong utility, additive difference, and diffusion) with machine learning methods (i.e., multinomial logistic regression, generalized additive model, twinned neural network, and Gaussian process). This study specifically focuses on the discrete two-alternative forced choice paradigm, where a choice must be made between two mutually exclusive alternatives [13]. Since the psychological process of how individuals make choices cannot be observed, this choice-making process must be inferred from their behavior [14]. There are many choice rules varying in functional forms (e.g., linear, monotonically increasing, or time-bound) and complexity [15] in decision science research. Machine learning methods have demonstrated their ability to identify complex nonlinear patterns in high-dimensional data [16]. Therefore, machine learning models must be able to discern which of these choice rules best approximates individuals' behavior based on a small sample of structurally similar choices.

This study compares the use of parametric, semi-parametric, and non-parametric machine learning applications in discrete choice modeling. These models must account for heterogeneity across individuals' choice rules, the process by which they select between alternatives [17–19]. The degree to which four commonly used machine learning models are suited for learning individual behavioral choice rules, when choice preferences are functionally complex, individuals are heterogeneous, and choice sets are limited, is examined. By testing these models across different choice rules and different choice-making contexts, this work aims to establish baseline performance metrics for these models. This approach enables the evaluation of each machine learning model's capacity for uncovering underlying choice rules without relying on restrictive assumptions, and examines the robustness of these models using an end-to-end controlled data-generating process. The main contribution of this work includes a demonstration of how semi- and non-parametric models can improve the learning of discrete choice decision rules under different decision contexts. The findings in this study contribute to both the theoretical understanding of discrete choice behavior and the practical application of machine learning tools in behavioral modeling, offering insights into how individuals navigate complex decision environments and informing how model selection can be done in different decision-making contexts.

The remainder of this paper includes a literature review, methodological overview of the choice rule data generation process and the machine learning models, experimental simulation parameters and results, and a case study of model performance on energy policy preferences. This is followed by a discussion on the benefits and trade-offs of model selection, limitations of this work, and future directions.

## 2. Related work

While there are many different framings for understand individual decision-making, discrete choice is a common preference elicitation method, for it has a variety of applications in policy domains, such as healthcare [20–23], transportation [1,2,24–26], and energy system preferences [7,18,27,28]. Throughout the literature, various models have been proposed on how an individual could decide between two or more discrete alternatives. Luce and Suppes proposed a random utility model, where while a decision-maker is assumed to have a self-defined utility function that allows them to choose the alternative with the greatest utility, the observer to the decision-making can only describe the decision-maker's preference up to some probability of choosing one alternative due to there being a random unobservable component in their decision-making process [29,30]. Marschak related the observation of a decision-maker's probabilistic choice behavior to the decision-maker's underlying preference in the form of a utility model [31]. Both Marschak and Luce and Suppes examined the Fechnerian model of utility [32] (also called strong utility model by Luce and Suppes [30]), where an individual's choice probability is determined by a difference in their utility function $u$, transformed by a strictly increasing continuous function $F$, conditional on the fact that their utility function is also strictly increasing continuous function. Coombs proposed the ideal point model [33] as a challenge to the Fechnerian idea, such that in addition to choices where utility has to be maximized (highest miles per gallon) or minimized (lowest price), there exist choices where the best option is an ideal intermediate option (the size of a car). Following this line of thinking, Tversky proposed the lexicographic order and the lexicographic semiorder [34], a decision-making rule where people make their choice by difference in attributes rank-ordered by importance. The lexicographic semiorder has the addendum that the difference in attribute must exceed a noticeable psychological difference to the individual. Busemeyer et al. added to the theories of decision-making by introducing the Decision Field Theory [35,36] and the Multi-alternative Decision Field Theory [37], where a time-bound deliberation component is included in the process, and the desirability of alternatives in a choice can change with time. Trueblood et al. extended upon the Multi-alternative Decision Field Theory by proposing a simplified Multi-attribute Linear Ballistic Accumulator model [38–41], where how much a decision-maker favors one option is determined by a weighted difference of subjectively-mapped objective attributes (such as price or quantity). Collectively, these mathematical formulations used to hypothesize how an individual may make a discrete choice are known as choice rules.

Separate from the discrete choice rules, which intend to model the psychological process of an individual's discrete decision-making paradigm, discrete choice modeling estimates and predicts how an individual may pick between two or more discrete alternatives based on the choices they have made. These models are classified into parametric models that assume a prior functional form and use a finite set of parameters, non-parametric models that have no prior assumption or set of fixed parameters [42], or semi-parametric models that make some, but not all, prior assumptions. This mode of analysis was pioneered by McFadden [43], who showed the relationship between utility maximization and probabilistic choice modeling. McFadden proved that a decision-maker who chooses according to the maximum of the random utilities of the alternatives can be modeled using a multinomial logistic function [11,43,44]. There are many parametric and semi-parametric machine learning models that extends upon the linear form of the multinomial logit, including the generalized additive models (GAMs) [11,45,46], generalized linear model elastic nets (glmnets) [47,48], and twinned neural network [49–51]. Advancements in machine learning also brought about the use of non-parametric models for discrete choice modeling, such as decision trees [52–54], random forests [54–57], and Gaussian process choice models [16,58–60]. Additionally, there have been developments of more psychologically-motivated machine learning models, such as the best estimate and simulation tools (BEAST) [61], the psychological forest [62,63], and the mixture of differentiable decision theories [64].

Within the literature, there is a gap between modeling an individual's psychological discrete choice-making process and estimating an individual's discrete choice preference, as they are generally different modeling paradigms. There are no studies that contribute to a fundamental evaluation of how discrete choice modeling paradigms perform in the context of the various psychologically motivated discrete choice rules. Thus, it is the goal of this study to provide a thorough benchmark assessment of parametric and non-parametric machine learning methods applied to discrete choice learning.

## 3. Method

We evaluate the ability of four machine learning methods to learn individual choice rules drawn from decision science using Monte Carlo simulations. The five choice rules that was selected represent different preference structures, each having distinct functional forms. Fig. 1 show this work's approach, adapted from Morris et al. [65]. Here, a choice rule is the mathematical approximation of an individual's unobservable psychological choice-making process. What can be observed is the choice an individual makes when they select one of two alternative choice options in a choice set, where both alternatives are characterized by a set of attributes that describe each option. The machine learning methods are also evaluated in a case study using the discrete energy policy preference data from [18].

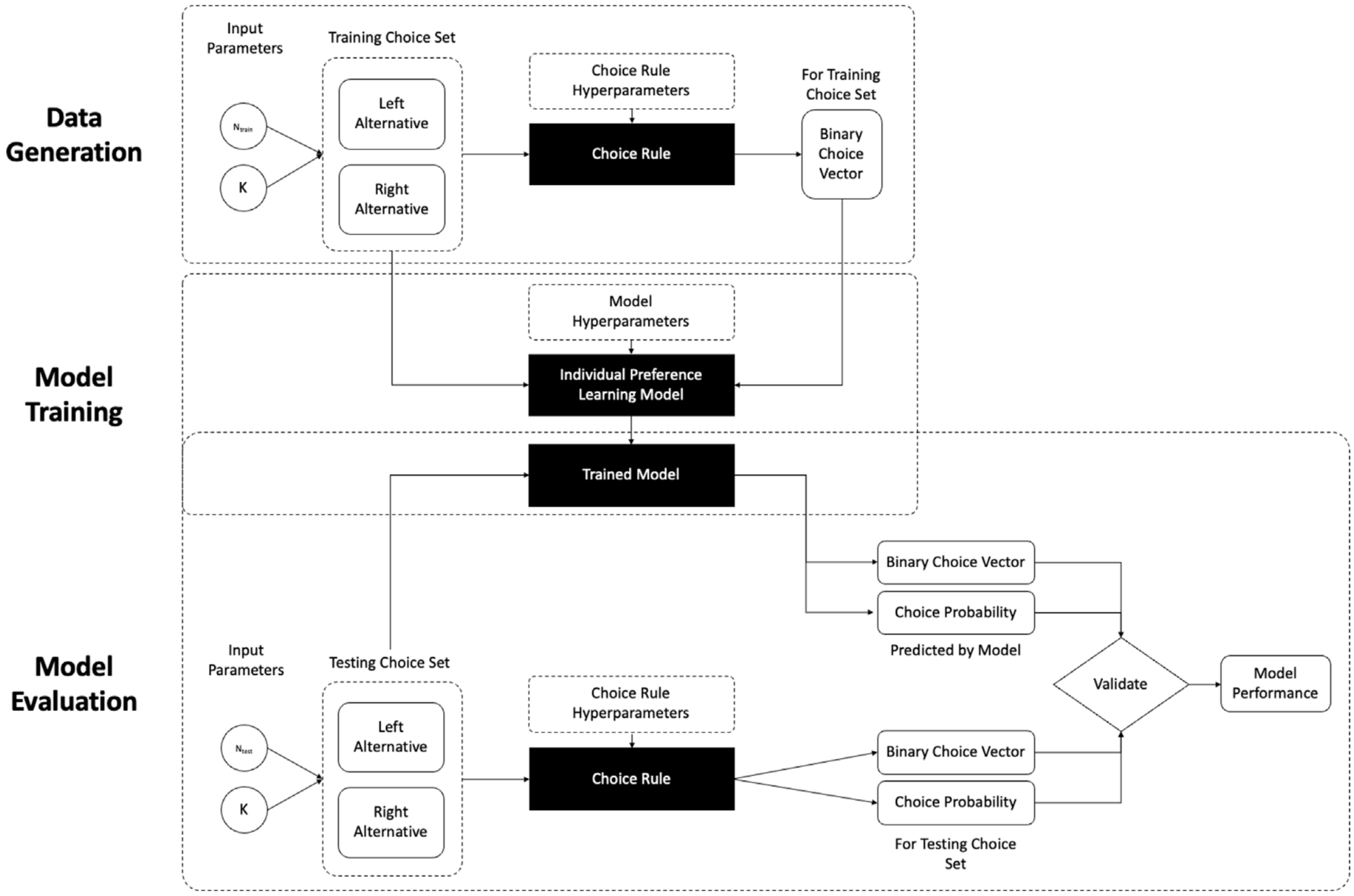

**Fig. 1.** Overview of the simulation process.

There are three phases to the simulation: data generation, model training, and model evaluation. The data-generating process creates two input vectors and one output vector for each choice rule. The input vectors are two-alternative choice sets generated by randomly sampling attribute levels from a uniform distribution for each pair of alternatives in each choice set. A binary vector of outputs is determined by the choice rule, which selects a single alternative from each choice set. The data-generation process is used to create input and output vectors of varying sizes for the training and testing choice sets. Across the experiments, the size of the training choice sets was varied, but the test choice sets was kept consistent ($N_{test} = 1500$ test choice sets). In the model training phase, the input vectors (attributes of the two alternatives in each choice set) and output vector (the selected alternative for each choice set as determined by the choice rule) are analyzed by each preference learning model. The model attempts to learn the choice rule that maps the input vectors to the output vector. In the model-evaluation phase, the trained model is used to predict the choice probabilities for each choice set in the test choice sets.

Each step in the simulation is grounded in the theoretical foundation of choice rule and machine learning literature. However, due to the stochasticity of the data generation process, designed to simulate decision heterogeneity even among individuals with the same choice rule, a Monte Carlo simulation is adopted. Each of the three phases is repeated for each combination of choice rule, model, and experimental hyperparameters during a 10-trial Monte Carlo simulation. The experimental parameters are outlined in Table 1.

### 3.1. Data-generating process

We created a data-generating process (Fig. 2) to simulate an individual making a two-alternative forced discrete choice using one of the five discrete choice rules. The data-generating process creates both the input choice sets vectors and a corresponding binary choice output vector based on the selected choice rule. The process requires the following hyperparameters: $N$ (number of choice sets), $K$ (number of attributes), and $\alpha$ (determinism coefficient, controlling the randomness of binary outputs using probabilistic choice rule).

The two-alternative (Left and Right) choice sets use attribute levels randomly sampled from a uniform distribution between 0 and 1. The five choice rules in Table 1 are applied to each choice set, leading to different output choice vectors. This random sampling allows each choice rule to take on multiple functional forms, such as varying the weights on a linear strong utility choice rule. The nature of random sampling may infrequently lead to attribute combinations that could not exist in the real world, which may lead to edge cases during model evaluation. This was accounted for through the Monte Carlo simulations.

### 3.2. Choice rules

A choice rule is a mathematical representation of how an individual selects among alternatives in a choice set. The five choice rules, common in decision science research, are selected due to their diverse functional forms. In this study, the strong utility and diffusion models use a Bernoulli random variable based on a latent probability, while the Additive Difference model uses an indicator function to select an alternative deterministically from each choice set. The following section outlines the formulation of each choice rule used in the study, with additional details present in Supplemental Information 1.

#### 3.2.1. Strong utility models

Strong utility choice rules probabilistically select from one alternative (L) relative to another (R) by using an increasing function of the difference in cardinal utility of the alternatives [30,43,66,67]. Given the choice sets as an input, Strong Utility choice rules output the probability of preferring the Left alternative over the Right alternative. The binary choice vector is generated using the choice probability and the Bernoulli random variable. This choice probability is computed using the logistic function in Eq. (1).

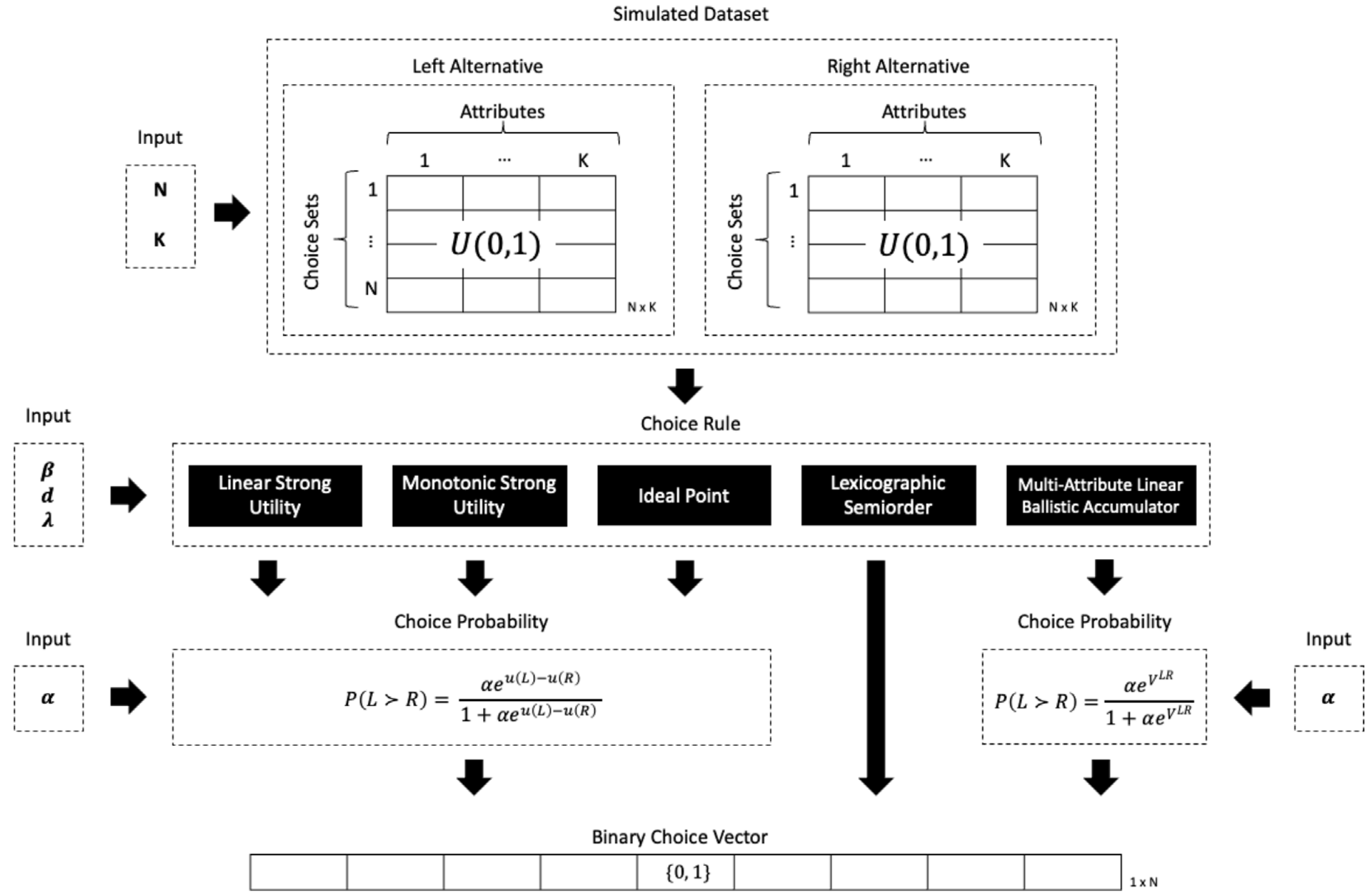

**Fig. 2.** Data-generating process.

**Table 1**
Summary of experimental data-generating parameter values.

| Experiment | Hyperparameter | Values | Constant | Constant values |
|---|---|---|---|---|
| Large choice set | $K$ (number of attributes) | 2 to 15 (increment by 1) | $N_{train}$ (number of training choice sets)<br>$N_{test}$ (number of testing choice sets)<br>$\alpha$ (randomness coefficient)<br>Attribute levels<br>Monte Carlo trials | 300<br>1500<br>10<br>$U(0,1)$<br>10 |
| Small choice set | $N_{train}$ (number of training choice sets) | 10 to 100 (increment by 10) | $K$ (number of attributes)<br>$N_{test}$ (number of testing choice sets)<br>$\alpha$ (randomness coefficient)<br>Attribute levels<br>Monte Carlo trials | 4<br>1500<br>10<br>$U(0,1)$<br>10 |
| Determinism | $\alpha$ (randomness coefficient) | 0 to 10 (increment by 1) | $N_{train}$ (number of training choice Sets)<br>$N_{test}$ (number of testing choice sets)<br>$K$ (number of attributes)<br>Attribute levels<br>Monte Carlo trials | 100<br>1500<br>4<br>$U(0,2)$<br>10 |

$$P(L \succ R) = \frac{\alpha e^{u(L)-u(R)}}{1+\alpha e^{u(L)-u(R)}} \tag{1}$$

Here, $\alpha$ is the determinism coefficient, representing an individual's certainty in the preference of their choice, and controls the noise in the logistic function. As the value of $\alpha$ increases, the choice probability approaches 0 or 1, making the alternative selection more deterministic. The following choice rules are the three strong utility choice rules studied here, where each has a different definition of the difference in cardinal utility $u(L) - u(R)$.

*Linear strong utility*

The linear strong utility choice rule weighs each attribute in both alternatives by the same weights. The difference in utility in Eq. (2) between the alternatives is taken as the weighted sum of the difference in the left and right attributes [15].

$$u(L) - u(R) = \sum_{k=1}^{K} \beta_k \left(x_k^L\right) - \beta_k \left(x_k^R\right) \tag{2}$$

$x_k^L$ and $x_k^R$ are attribute levels for the $k$th attribute for the left and right alternatives, respectively, and $\beta_k$ is the corresponding weight for the attribute, for up to $K$ attributes. The $\beta_k$ weights represent the relative importance of each attribute when making the choice, where a more important attribute has a higher weight. They are hyperparameters randomly uniformly generated between 0 and 1 during the data-generating process.

*Monotonic strong utility*

The monotonic strong utility choice rule computes the difference in utility between the left and right alternatives through a weighted sum of the difference between the monotonically transformed attribute levels [15]. The difference in utility is seen in Eq. (3).

$$u(L) - u(R) = \sum_{k=1}^{K} \beta_k f\left(x_k^L\right) - \beta_k f\left(x_k^R\right) \tag{3}$$

Here, $f$ can be any monotonic function, and the hyperbolic tangent function tanh was employed. This monotonic transformation reduces large attribute levels to a smaller range while preserving the order of the levels, so attributes with larger levels do not dominate the choice. Similar to linear strong utility, monotonic strong utility uses randomly uniformly generated weight values for $\beta_k$.

*Ideal point*

The ideal point choice rule assumes that psychologically, every decision maker has an ideal value for each attribute [33] and uses that value to make their selection. The ideal point choice rule computes the difference in utility as the weighted sum of the square difference between each alternative's attribute levels and the ideal values [15]. The difference in utility is calculated as in Eq. (4).

$$u(L) - u(R) = \sum_{k=1}^{K} \beta_k \left(x_k^L - d_k\right)^2 - \beta_k \left(x_k^R - d_k\right)^2 \quad (4)$$

Here, $d_k$ is the ideal value for the $k$th attribute. The data-generating process randomly generates both the attribute coefficient $\beta_k$, and the median attribute level is used as the ideal value $d_k$. The ideal values in the original model are random variables, but they are treated as constants in this work.

#### 3.2.2. Additive difference

Additive difference is a non-probabilistic choice rule that compares attributes sequentially, using a difference function $\phi_k$ [13,34]. The present research uses lexicographic semiorder as an example of additive difference choice rules.

*Lexicographic semiorder*

The lexicographic semiorder choice rule assumes that the decision maker ranks the attributes and then selects the alternative with the best value on the most important attribute [34]. lexicographic semiorder computes the binary choice vector directly with an indicator function $\mathbb{I}$ [15] in Eq. (5), returning either 1 (selecting the left alternative) or 0 (selecting the right alternative).

$$C(L \succ R) = \mathbb{I}\left[x_m^L - x_m^R > \epsilon_m\right] \quad (5)$$

$C(L \succ R)$ is the choice between the left and the right alternative. Attributes are ranked, and the decision is made on the $m$th-highest ranked attribute. When the $m$th attribute difference exceeds $\epsilon_m$, the Just Noticeable Difference, the Left alternative is selected. If none of the attribute differences exceed the Just Noticeable Difference, the choice is made randomly. The data-generating process randomly ranks the available attributes, using a Just Noticeable Difference of 0.2.

#### 3.2.3. Diffusion

Diffusion choice rules assume that decisions are made through a noisy process in which information is accumulated over time, from the starting point towards the response criteria [68]. This study uses the MLBA as an example of a diffusion choice rule.

*Multiattribute linear ballistic accumulator (MLBA)*

The MLBA is a diffusion model that includes a subjective mapping and a weighting function in the calculation of the valence difference [40]. The valence difference Eq. (6) indicates how much one alternative is favored over the other, based on how much attention the decision-maker pays to the difference in subjective valuations.

$$V^{LR} = \sum_{k=1}^{K} w_k^{LR} \left(U\left(x_k^L\right) - U\left(x_k^R\right)\right) \quad (6)$$

The valence difference is the sum of the weighted differences for $K$ subjectively valuated attributes. The subjective mapping function $U$ transforms objective attribute quantities $x_k^L$ and $x_k^R$ into psychological magnitudes. This study uses the natural log as that subjective mapping function. The weighting function $w_k^{LR}$ is seen in Eq. (7).

$$w_k^{LR} = e^{-\lambda\left|U\left(x_k^L\right) - U\left(x_k^R\right)\right|} \quad (7)$$

The hyperparameter $\lambda$ determines the weight based on the difference in the psychological magnitude of the subjectively mapped attributes. This study uses the valence difference to compute the choice probability [15], applying the modified logistic function in Eq. (8).

$$\mathrm{P}(\mathrm{L} \succ \mathrm{R}) = \frac{\alpha \mathrm{e}^{\mathrm{V^{LR}}}}{1 + \alpha \mathrm{e}^{\mathrm{V^{LR}}}} \quad (8)$$

The choice probability is determined by the valence difference and the determinism coefficient $\alpha$. The binary choice vector is generated using the choice probability.

### 3.3. Machine learning models

This study compares four machine learning methods, being the multinomial logistic regression, generalized additive model, twinned neural network, and Gaussian process. Among these models, the Gaussian process is non-parametric, the generalized additive model is semi-parametric, and the other methods are parametric. These four methods are assessed for their ability to recover an individual's preferences across the five choice rules. See Supplemental Information 2 for detailed implementation on all models.

#### 3.3.1. Multinomial logistic regression (MNL)

Multinomial logistic regression (MNL) models are one of the oldest and most widely used discrete choice modeling paradigms [43], where if the random component of utility has an assumed independent Gumbel Distribution, then a decision-maker who chooses to maximize their random utility can have their choice probability represented as a multinomial logistic function. That choice probability for two alternatives takes the form of Eq. (9) [69]:

$$P\left(y_n^L = 1 | \beta\right) = \frac{e^{\beta^T x_n^L}}{e^{\beta^T x_n^L} + e^{\beta^T x_n^R}} \quad (9)$$

This function indicates the probability of choosing the left alternative (denoted as 1) as the utility of the left alternative over the sum of utilities of both alternatives, for the $n$th choice set. The utility of each alternative is the linear transformation of the attribute levels using a vector of coefficients $\beta$. The regression derives the $\beta$ coefficient terms through maximum likelihood estimation. Due to the assumption of linearity for the multinomial logistic function in (9), MNL is considered a parametric machine learning method of recovering an individual's preference.

#### 3.3.2. Generalized additive model (GAM)

The generalized additive model (GAM) [70] extends linear models, such as MNL, by using a linear predictor that links a univariate response variable to predictor variables. With binary discrete choices, a GAM is constructed by first performing the basis expansion on a single attribute [69], creating two basis expansion matrices for each alternative. The difference in utility equals the difference in the basis expansion matrices multiplied by a vector of unknown parameters. The unknown parameters are approximated using penalized maximum likelihood estimation, iteratively updated using the Newton–Raphson method [46]. GAM is considered semi-parametric because while the model is additive from any number of smooth differentiable univariate functions estimated from data, GAM is estimated using a parametric penalized maximum likelihood estimation.

#### 3.3.3. Twinned neural network (TNN)

The twinned neural network (TNN) [49,51,71] is an architectural variant of a neural network, a class of ML model capable of extracting underlying patterns from complex, multidimensional data. TNN performs nonlinear parametric estimation using its multiple layers associated with nonlinear activation functions with a specified functional form. The alternatives are separated into two input layers and fed into the TNN. Each alternative is transformed using three hidden layers, where the alternatives share the same weights. Two identical encodings are used to learn the data patterns of the Left and Right alternatives separately, to preserve permutation invariance, where switching the order of the alternatives should not affect the psychology of the decision. The difference across attributes in the transformed alternatives is taken during a differencing layer. Subjective values of the differenced alternatives are calculated in subsequent layers, and a binary choice is calculated using a logistic function. A hyperparameter tuning was performed for the TNN using a grid search method implemented through the Talos package [72]. The hyperparameters considered include the number of nodes and the activation functions for each hidden layer, the batch size, the number of epochs, the optimizer, and the loss function.

#### 3.3.4. Gaussian process (GP)

Gaussian process (GP) is a type of non-parametric regression model that defines a probabilistic distribution over functions, with the subsequent inference taking place within this space of functions [60]. This study uses a covariance kernel $K$ to capture the discrete choice process of selecting the left alternative over the right alternative [42,59]. The difference in utility between the left alternative ($\boldsymbol{X}^{\boldsymbol{L}}$) and the right alternative ($\boldsymbol{X}^{\boldsymbol{R}}$) can be approximated using a zero-mean GP Eq. (10).

$$u\left(\boldsymbol{X}^{\boldsymbol{L}}\right) - u\left(\boldsymbol{X}^{\boldsymbol{R}}\right) \sim GP(0, K) \tag{10}$$

The covariance kernel $K$ assesses the similarity between the left and right alternatives of two choice sets. Choice probability is predicted using the standard Gaussian CDF applied to the difference in utility. Unlike TNN, the GP model as defined by this kernel is not permutation invariant. GP uses the number of iterations and the number of chains as training hyperparameters. In this study, 50 iterations and 1 chain are used to train the GP model, and the hyperparameter tuning is performed using a grid search.

### 3.4. Experimental design

We conduct a series of simulations as a model performance assessment study. The simulations consist of three Monte Carlo experiments, selected to evaluate each model's capacity for recovering choice rules in the presence of large amounts of information presented to the decision-maker, limited responses from the decision-maker, or varying uncertainty in preference between alternatives. The **large choice sets** experiment assesses the effect of expanding the number of attributes on each preference learning model's capacity for recovering the choice rules. The **small choice sets** experiment assesses each model's performance with a limited set of training data. The **determinism** experiment assesses model performance when probabilistic choice rules become more deterministic. The Monte Carlo simulation repeats each experiment 10 times, addressing the concern that random sampling may lead to unrealistic attribute combinations in the generated choice sets.

### 3.5. Accuracy metrics

This study reports model fit using the Bayesian Information Criterion (BIC). The Bayesian Information Criterion (BIC) is a metric used for model selection that penalizes for overfitting by adding more modeling parameters [73]. It is defined in Eq. (11) as:

$$\mathrm{BIC} = k \ln(n) - 2 \ln\left(\hat{L}\right) \tag{11}$$

Here, $k$ is the number of parameters estimated by the model, and is translated into the number of attributes for each choice set. The number of choice sets is represented by $n$. $\hat{L}$ is the maximized value of the likelihood function of the model. When selecting from several models, the model with the lowest BIC value is the most preferable.

The benefit of using BIC for model selection is that it strongly penalizes for the complexity of the model to prevent overfitting, and can measure the efficiency of model prediction. The major limitation of BIC includes that the calculation for BIC, as represented in Eq. (11), is only valid if $n \gg k$, and that BIC cannot handle high-dimensional datasets [74].

For further robustness, five additional accuracy metrics was used to evaluate the machine learning model's ability to recover the choice rules: average negative log-likelihood, normalized likelihood, Brier score, average standard error, and binary false classification rate. Decomposed Brier scores are used to visualize the model's predictive calibration. Each metric compares the model's probabilistic and binary predictions against the observed outcome from the choice rule. See Supplemental Information 3 for more details and limitations of each metric. Different metrics are used to accommodate the uncontrolled variation due to the random data-generating process. See Supplemental Information 4 for a discussion on sources of uncertainty and a proposed method to perform significance testing.

## 4. Results

This section describes the large choice sets, small choice sets, and determinism experiments, where the experimentation parameters are summarized in Table 1. Fig. 3 summarizes the model selection process by outlining which of the four machine learning models is the best performing for capturing which choice rule under the three decision contexts as simulated in the experiments.

### 4.1. Large choice sets experiment

The large choice sets experiment evaluates each model's predictive performance for different choice rules as the complexity of choice sets increases from 2 to 15 attributes. Here, four machine learning models was trained using 300 training choice sets, and evaluated the models using 1500 testing choice sets. A Monte Carlo simulation is repeated for this experiment 10 times.

Fig. 4 reports the model performance using the Bayesian Information Criterion (BIC) for each choice rule-model pair. It demonstrates that some models perform better than others at learning each choice rule, but no model is superior for learning all choice rules. GAM performed the best for linear strong utility (BIC value of 690.6) and MLBA choice rules (851.2), GP performed the best for ideal point (1723.8) and monotonic strong utility choice rules (1787.1), and TNN performed best for lexicographic semiorder (389.9). A lower value of BIC indicates better model fit. Thus, the best-performing model for the choice rule is one with the lowest BIC.

Fig. 5 examines the change in model performance as the number of attributes increases, where each subplot represents a single choice rule. Linear (Fig. 5a) and monotonic strong utility (Fig. 5b) displayed 24%–41% and 29%–37% decreases in BIC, respectively, as the number of attributes increased. For the ideal point rule (Fig. 5c), GAM and GP performed 10%–11% better as the number of attributes increased, while TNN and MNL performed 6% worse as the number of attributes increased. Unlike other rules, the error on lexicographic semiorder (Fig. 5d) and MLBA (Fig. 5e) increased by 49%–92% and 103%–120% respectively, with the number of attributes. See Table 2 for model-specific values. The Monte Carlo uncertainty for models MNL and TNN increased for the linear strong utility and monotonic strong utility rules, remained unchanged for ideal point and lexicographic semiorder, and decreased for MLBA. In contrast, the uncertainty on GAM and GP remained unchanged for ideal point and lexicographic semiorder,

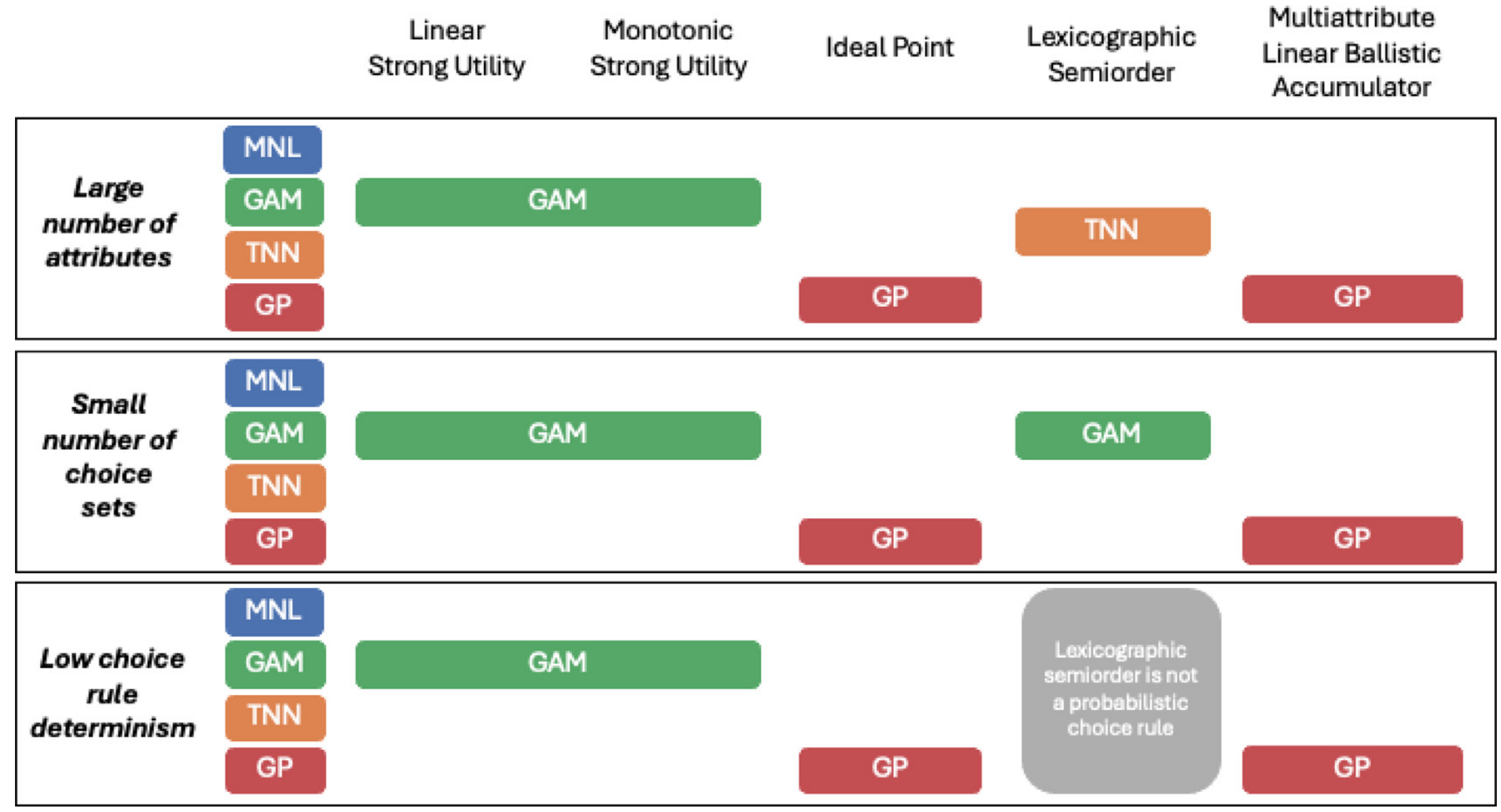

**Fig. 3.** Best performing model for each choice rule under simulated decision-making contexts.

**Table 2**
Percent change Bayesian information criterion for large choice sets experiment (Fig. 5).

| | MNL | GAM | TNN | GP |
|---|---|---|---|---|
| Linear strong utility | −40% | −41% | −40% | −24% |
| Monotonic strong utility | −36% | −37% | −36% | −29% |
| Ideal point | +6% | −10% | +6% | −11% |
| Lexicographic semiorder | +65% | +69% | +92% | +49% |
| MLBA | +103% | +116% | +116% | +120% |

Note: 2 and 15 attributes are compared. Negative (−) indicates a percent decrease, positive (+) indicates a percent increase, and asterisk (*) indicates a direct change from 0.

**Table 3**
Percent change in Bayesian information criterion for small choice sets experiment (Fig. 6).

| | MNL | GAM | TNN | GP |
|---|---|---|---|---|
| Linear strong utility | −96% | −60% | −68% | −48% |
| Monotonic strong utility | −51% | −56% | −62% | −45% |
| Ideal point | −72% | −69% | −62% | −6% |
| Lexicographic semiorder | −68% | −54% | −73% | −62% |
| MLBA | −72% | −58% | −57% | −20% |

Note: 10 and 100 training choice sets were compared. Negative (−) indicates a percent decrease, positive (+) indicates a percent increase, and asterisk (*) indicates a direct change from 0.

and decreased for other rules. Uncertainty did not decrease for the lexicographic semiorder rule, which is potentially due to the choice rule's non-probabilistic nature. The same trend is observed from average negative log-likelihood and binary false classification rate metrics (see Supplemental Information 5), but average standard error increased with the number of attributes for the Strong Utility choice rules. As the number of attributes increases, models' forecasting error for strong utility choice rules is increasing at a granular probabilistic level, but the probabilistic forecast error relative to the observed binary outcome decreased (see Supplemental Information 5.4 for more discussion).

Thus, in this scenario, with an abundance of choice sets for training, as the number of attributes increases, Monte Carlo uncertainty in model performance decreases for the probabilistic choice rules. When measuring against only the observable outcome (binary choice from choice rule), the increase in the number of attributes leads to a decrease in error for the strong utility rules, and an increase in error for lexicographic semiorder and MLBA. The semi-parametric GAM model predicts the best for linear strong utility and MLBA, the non-parametric GP model predicts the best for ideal point and monotonic strong utility, and the parametric TNN model performs best for lexicographic semiorder. See Supplemental Information 5 for performance summaries using other metrics (Brier score, normalized likelihood, binary false classification rate, and average standard error).

### *4.2. Small choice sets experiment*

The small choice sets experiment evaluates each model's performance when there are just 10 to 100 training choice sets, each with 4 attributes. The purpose of this experiment is to examine the impact of a limited training choice sets on each model's predictive performance.

In Fig. 6, GAM predicted the best for linear strong utility (BIC value of 1138.0), monotonic strong utility (1298.6), and lexicographic semiorder (514.2), and GP performs best for ideal point (2056.3) and MLBA (1646.8). See Supplemental Information 6 for additional performance summaries using other metrics.

Fig. 7 visualizes the impact of limiting training choice sets on model performance. As the training choice sets size increases, the Monte Carlo uncertainty on the BIC decreases for all models across all choice rules, except MLBA. In general, the average model predictive error also decreases as more training choice sets become available. BIC decreased by 48%–96% for linear strong utility (Fig. 7a), 45%–62% for monotonic strong utility (Fig. 7b), 5%–72% for ideal point (Fig. 7c), 54%–73% for lexicographic semiorder (Fig. 7d), and 20%–72% for MLBA (Fig. 7e). See Table 3 for model-specific values. This trend is also observed when using all other metrics (see Supplemental Information 6).

Increasing the training choice sets from 10 to 100 reduces both Monte Carlo uncertainty and model prediction error. When there is a limited amount of discrete choice sets available to infer an individual's preference, the semi-parametric GAM model predicts linear strong utility, monotonic strong utility, and lexicographic semiorder rules the best, while the non-parametric GP model predicts ideal point and MLBA the best.

#### *4.2.1. Determinism experiment*

The determinism experiment increases the determinism coefficient $\alpha$ for all probabilistic choice rules from 0 (random) to 10 (deterministic). The determinism coefficient controls how probabilistic the strong utility and diffusion choice rules are, where a higher coefficient value reflects the idea that the individual is making a choice with a stronger preference, as opposed to choosing randomly with a low determinism coefficient.

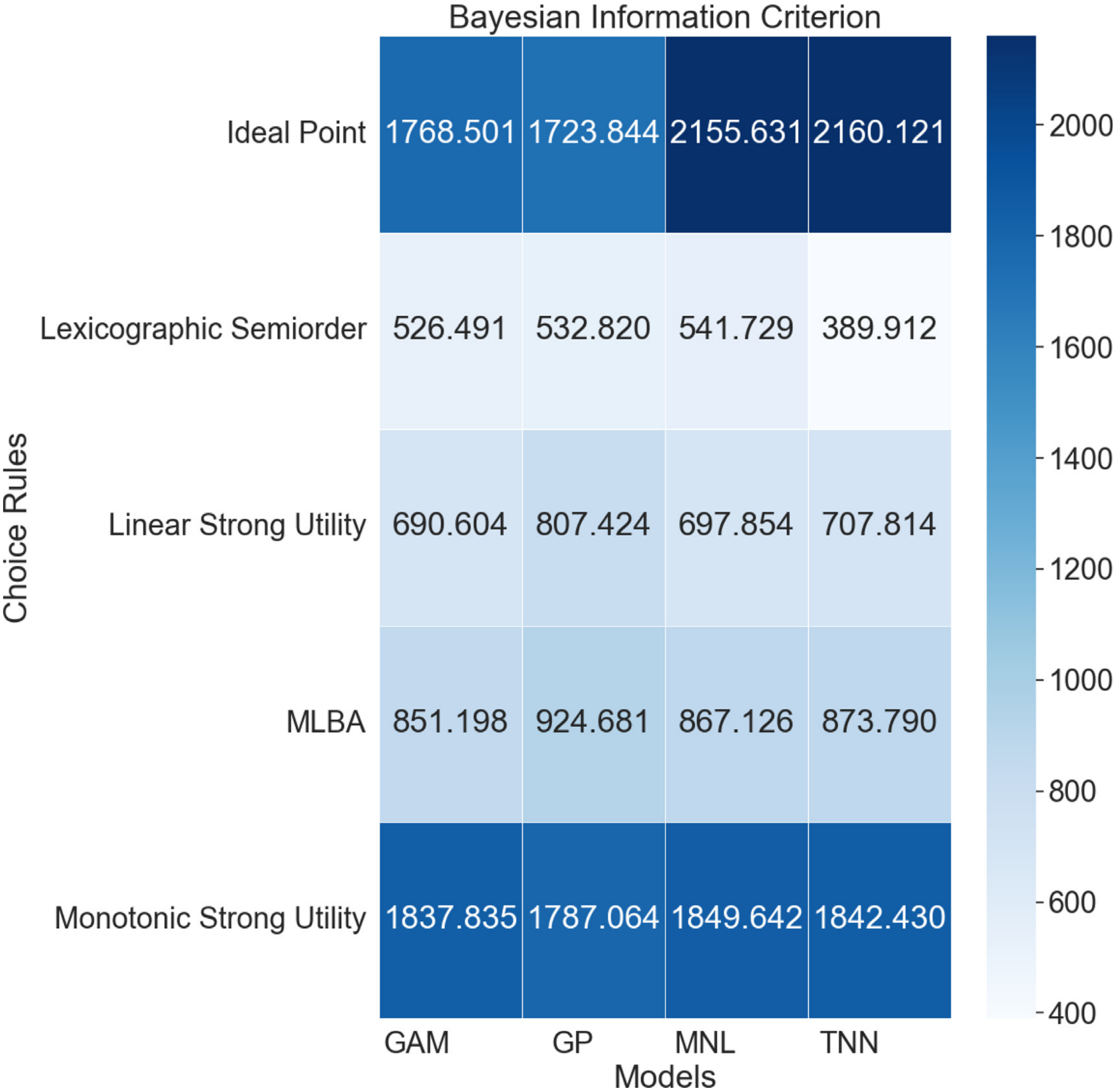

**Fig. 4.** Model performance summary for the large choice sets experiment with the Bayesian Information Criterion (BIC).
Note: Lower numbers (lighter colors) indicate better model fit. Errors are averaged across both Monte Carlo trials and the changing number of attributes.

The lexicographic semiorder is excluded from this analysis as it uses a non-probabilistic choice rule. The models are trained on $N_{train} = 100$ choice sets. The purpose of this experiment is to examine the impact of a progressively more deterministic choice rule on models' predictive performance.

In Fig. 8, GAM performed the best for linear and monotonic strong utility, with an average Bayesian Information Criterion (BIC) measurement of 1454.0 and 1598.6, respectively. GP performed the best for ideal point (BIC value of 2070.3) and MLBA (BIC value of 1732.8). See Supplemental Information 7 for performance summaries using other metrics.

In Fig. 9, BIC decreased by 48%–55% for linear strong utility (Fig. 9a), 43%–49% for monotonic strong utility (Fig. 9b), and 31%–34% for MLBA (Fig. 9d). For ideal point (Fig. 9c), BIC increased by 0.2%–0.4% for MNL and TNN, and decreased by 5%–7% for GAM and GP, observing the same split performance as the large choice sets experiment (Fig. 5). A decrease in BIC indicates better model fit. See Table 4 for model-specific values. The increase in determinism increased the Monte Carlo uncertainty of BIC. The same trend is observed with average negative log likelihood and binary false classification rate, but average standard error increases with the increase in determinism (see Supplemental Information 7 for discussion).

The models predict better when decision-makers use more deterministic versions of the linear strong utility, monotonic strong utility, and MLBA choice rules. However, that was not true for the ideal point choice rule, which this study interprets as reflecting the attribute levels used (see Supplemental Information 7.5 for discussion). Overall, when there is uncertainty in how deterministic an individual's choice preference is, the semi-parametric GAM model performs the best for linear and monotonic strong utilities due to lower BIC values, and both the semi-parametric GAM model and the non-parametric GP model perform comparably for ideal point and MLBA.

### 4.3. Computational complexity

Fig. 10 showed the average model training time in seconds for each of the four models in the three simulation scenarios. All models were trained on a 2021 Apple M1 chip. Across all choice rules and scenarios, GP is uniquely the most computationally expensive model out of the four tested, and MNL is by far the least computationally expensive. This is to be expected, as model training time is correlated with model complexity, and the non-parametric GP model is an iterative and nonlinear model. However, training time is also dependent on the computing hardware. While GP is a high-performing model, it would

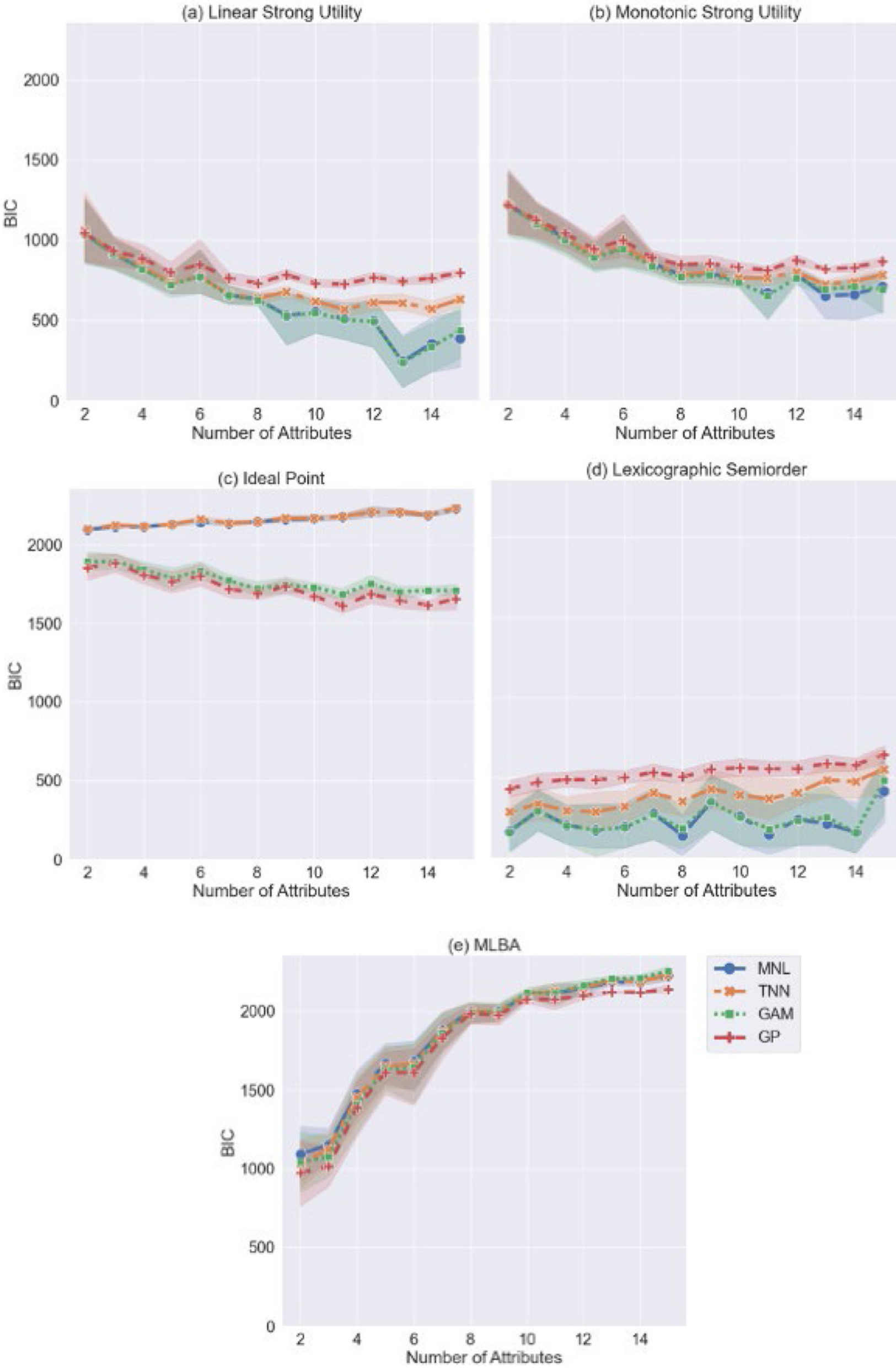

**Fig. 5.** Bayesian information criterion for the large choice sets experiment.
Note: The scores are averaged across all Monte Carlo trials. The shaded region displays two standard deviations within all brier scores across the Monte Carlo trials for each number of attributes. Increasing the number of choice set attributes affects model prediction differently, depending on the choice rule.

**Table 4**
Percent change in Bayesian information criterion for the determinism experiment (Fig. 9).

| | MNL | GAM | TNN | GP |
|---|---|---|---|---|
| Linear strong utility | −54% | −55% | −54% | −48% |
| Monotonic strong utility | −47% | −49% | −46% | −43% |
| Ideal point | 0% | −7% | 0% | −5% |
| MLBA | −32% | −34% | −31% | −31% |

Note: Determinism coefficients of 0 and 10 were compared. Negative (−) indicates a percent decrease, positive (+) indicates a percent increase, and asterisk (*) indicates a direct change from 0.

be beneficial to select GAM or TNN if computation time is a concern in situations with a lack of advanced computing hardware, or application contexts with large choice sets and attributes.

### 4.4. Case study

The performance of each model is evaluated against a real public policy dataset on discrete energy policy preferences [18,28]. The dataset consists of 822 survey participants, where each survey participant compare between two sets of energy policy portfolios. Each participant was shown $N = 16$ two-alternative 4-attribute forced choice sets, where they chose their preferred energy policy portfolio alternative per choice set. The attributes characterizing the alternatives are electricity portfolio, monthly electricity bill, climate change-related emissions, and health-related pollution. Model performance is assessed per survey participant, where the preference learning model trained on $N_{train} = 10$ choice sets, and the predictive accuracy of each model is validated on $N_{test} = 6$ choice sets.

The model fit is assessed using BIC, averaged across all survey participants in the discrete choice experiment. Across the sample, TNN

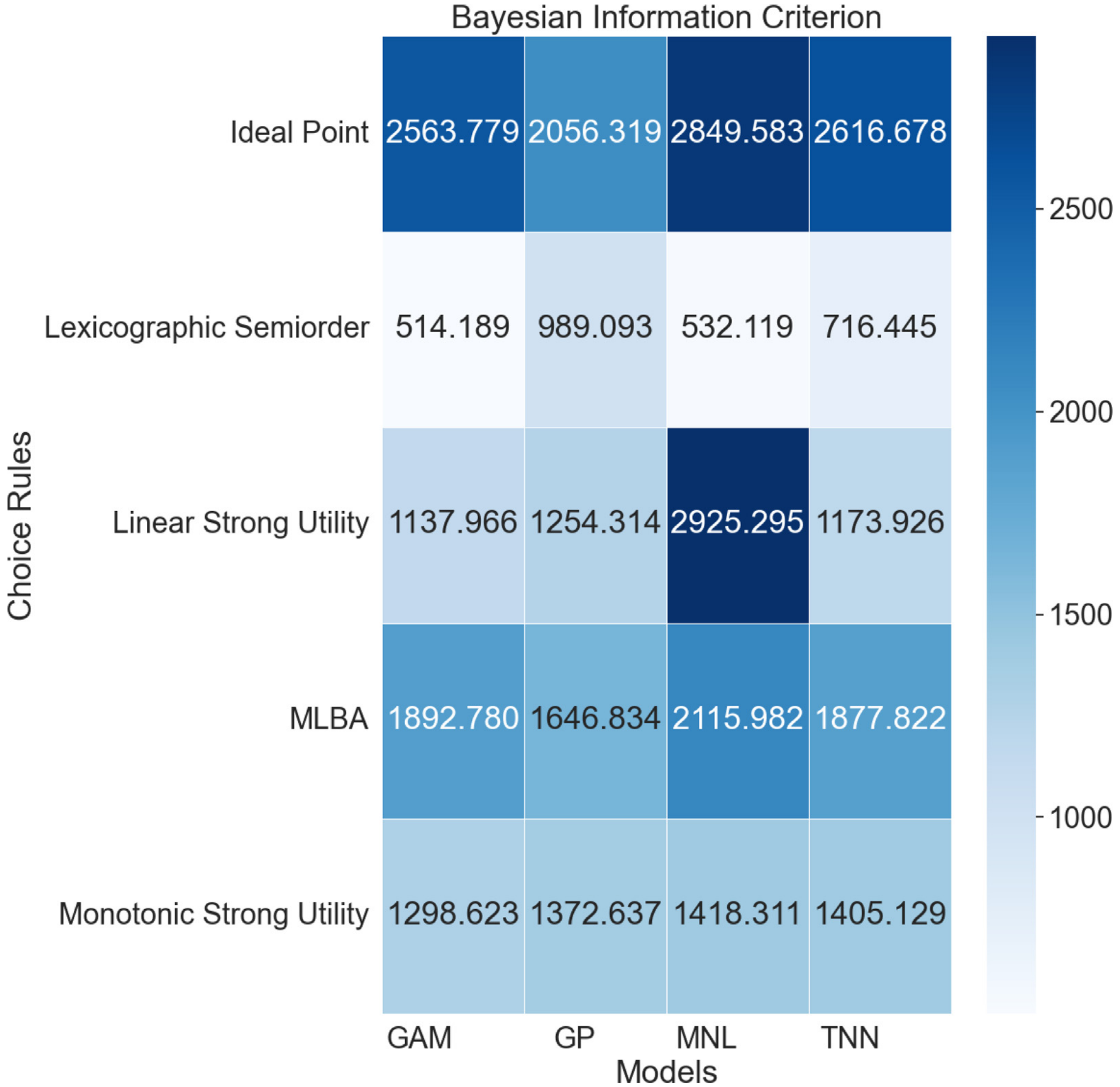

**Fig. 6.** Model performance summary for the small choice sets experiment with the Bayesian Information Criterion (BIC).
Note: Lower numbers (lighter colors) indicates better model fit. Errors are averaged across both Monte Carlo trials and the changing number of attributes.

and GP have the best model fit, with the lowest BIC values at 13.351 and 13.433, respectively. GAM has the worst model fit with a BIC value of 20.115, and MNL falls in the middle with a BIC value of 17.491. Fig. 11 visualizes how many of the testing choice sets were predicted correctly per individual. Within the same sample population, GAM most likely predicted 3 or 4 choice sets correctly, GP most likely predicted 4 or 5 correctly, and MNL and TNN most likely predicted 5 or 6 correctly. While MNL demonstrated comparable predictive performance with TNN, the MNL model was also significantly overfitting in most cases. 78% of the trained model, or 642 of the 822 samples, had at least one predictive probability of 0 or 1, resulting in a null BIC value. This was also observed for GAM, the worst-performing model in this case, with 30 samples resulting in a null BIC value.

This experiment demonstrated that all preference learning models can be trained and used to predict with real discrete choice sets. Assessing model performance using real survey data provides an understanding of model usability, should it be deployed as a policy decision aid to stakeholders.

## 5. Discussion

Machine learning models can non-parametrically capture an individual's discrete choice preferences by learning an individual's behavioral choice rules without making assumptions about how they make choices. However, preference learning is difficult due to individual heterogeneity, limited availability of choice sets, and malleability of choice rules. This study benchmarks the performance of four parametric and non-parametric machine learning models' capacity for learning discrete choice rules under these constraints. Four machine learning models was evaluated against five choice rules selected from decision science literature, in a framework that simulates (1) a large set of decision alternatives, (2) limited choice sets, and (3) varying determinism of preference. It was found that predictive performance improves with an increase in the number of training choice sets and choice rule determinism, but performance differs by choice rule when the number of attributes increases. Generally, GP learns the ideal point choice rule better than other models, but GAM is best at learning the linear strong utility choice rule. Overall, the ideal point choice rule is difficult to learn, as demonstrated by the high BIC values across all models in all experimental contexts. As the ideal value $d_k$ approaches 0.5, the maximum difference in utility becomes smaller, thus decreasing the strength of preference. Models in general have difficulties learning choice rules with weak choice preference, as weak preferences indicate a higher amount of noise. A detailed discussion on the uncertainties behind the ideal point choice rule can be found in Supplemental Information 4.

When many decision attributes are present, GAM also learns MLBA the best, GP learns monotonic strong utility the best, and TNN learns lexicographic semiorder the best. When there are limited choice sets present, or when there is more randomness in preference, GAM also learns monotonic strong utility and lexicographic semiorder the best, whereas GP learns MLBA the best. In this study, it was observed that no single machine learning model clearly dominates performance for all choice rules, but the semi-parametric GAM model and the non-parametric GP model generally outperform the parametric MNL and TNN models across the three examined contexts. The theoretical flexibility of the semi- and non-parametric models allows the model to better learn and adapt to the nonlinear and increasingly complex choice rules. This is a benefit when compared to the parametric models with their predefined parameter set and limited flexibility. When applied to policy contexts involving complex decision-making, the semi- and non-parametric models would be more accommodating.

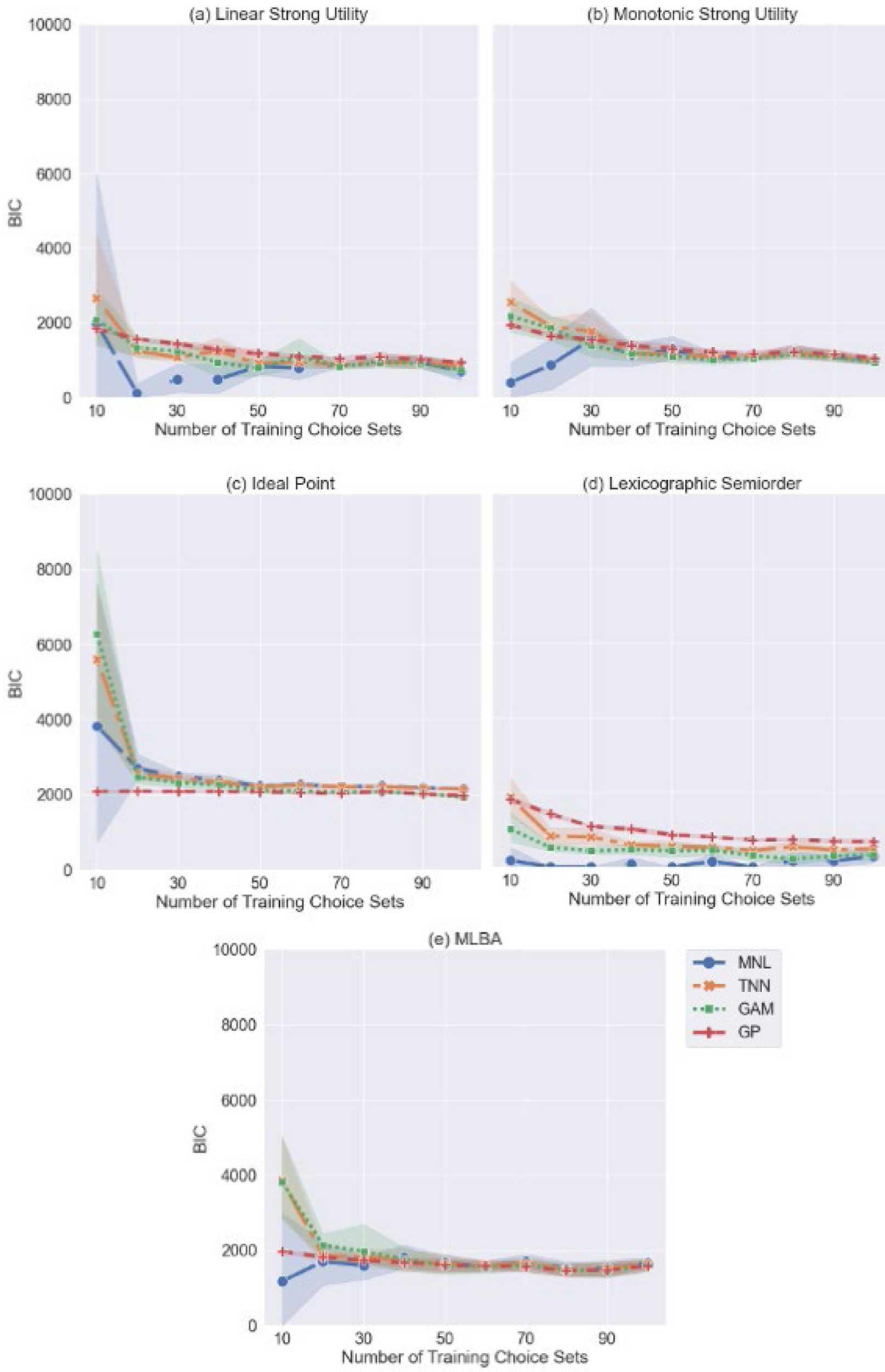

**Fig. 7.** Bayesian information criterion for the small choice sets experiment.
Note: The scores are averaged across all Monte Carlo trials. The shaded region displays two standard deviations within all brier scores across the Monte Carlo trials for each number of training choice sets. Increasing number of training choice sets decreases model prediction error.

While GAM and GP generally outperform MNL and TNN in the simulation, there are significant trade-offs between performance with linear interpretability and computation time. As a model, MNL is far more interpretable than other methods and has the lowest computational complexity, largely due to their linearity. However, as demonstrated by the case study, the MNL model significantly overfits in situations with low data availability. Despite the trade-offs, semi- and non-parametric models can be more flexibly applied across different application contexts.

### 5.1. Limitations

This research is limited by the point estimations used as summary error metrics to assess model performance. Point estimations succinctly summarize the performance of each model relative to the choice rule, allowing us to effectively compare model performances across choice rules. However, the point estimations also amalgamate the many sources of uncertainty present in the simulation, such as randomly sampled attribute levels, possibly creating infeasible combinations of choices, and the stochastically generated choice rule hyperparameters creating a choice rule that is not psychologically sound. See Supplemental Information 4 for a detailed discussion on sources of uncertainty present in this study. Significance testing on each choice rule-model pair in each experiment would untangle the impact of individual sources of uncertainty on the overall uncertainty of the point estimation and offer insight into the reliability of different accuracy metrics.

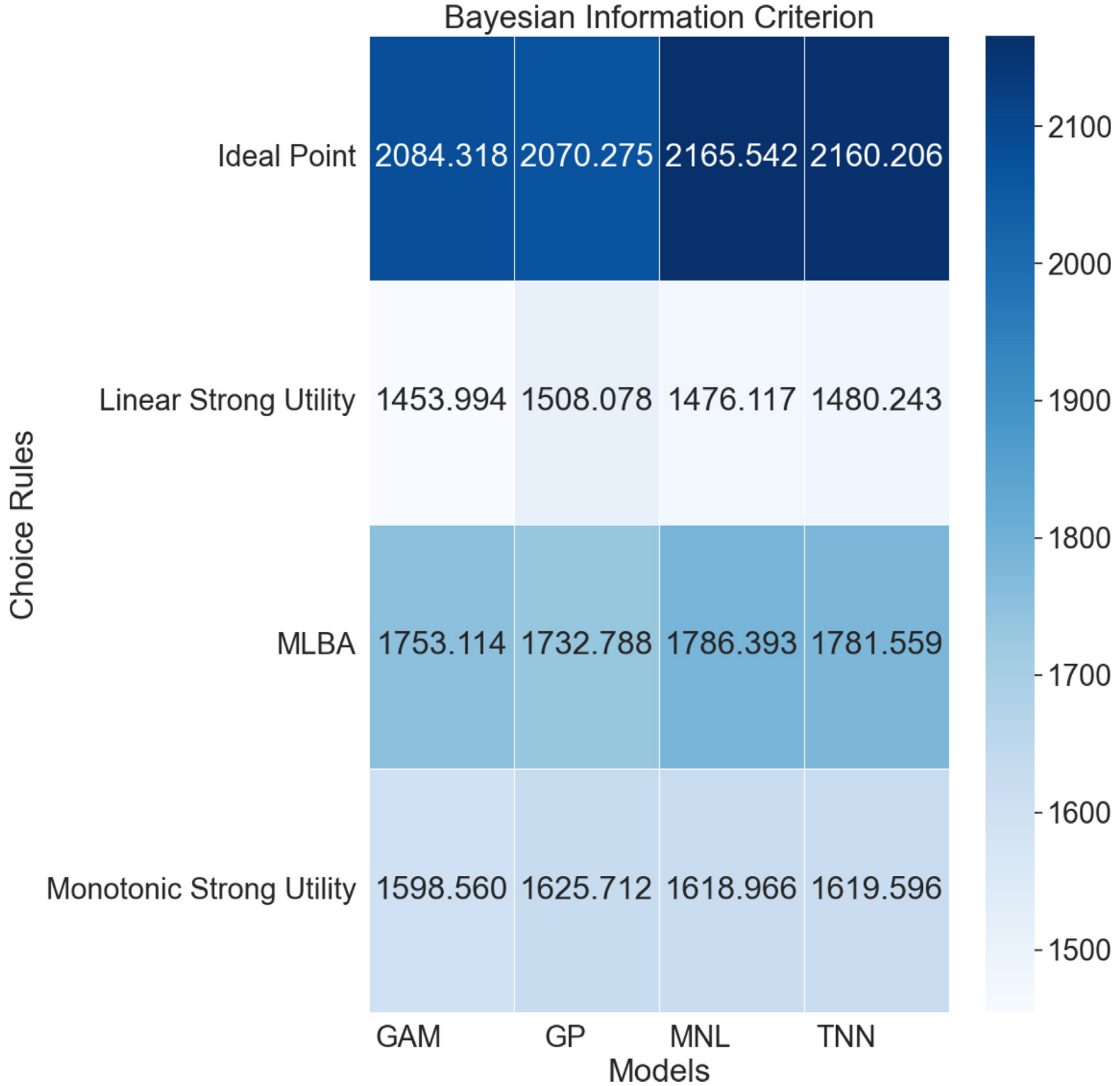

**Fig. 8.** Model performance summary for the determinism experiment with the Bayesian Information Criterion.
Note: Lower numbers (lighter colors) are more desirable. Errors are averaged across both Monte Carlo trials and the changing number of attributes.

We chose a Monte Carlo simulation method to capture the prediction uncertainty of each model, and the results demonstrated how prediction uncertainty changes due to the number of attributes, the number of training choice sets, and the determinism of the choice rule. However, a limited number of trials were performed due to the highly computationally expensive training. Despite the many accuracy measures used to fully assess the robustness of model prediction, certain metrics are unable to discern the best-performing model at the reported significant figures, or they disagree on the best-performing model.

Finally, the real discrete choice sets size per person is very small ($N_{train} = 10$), significantly limiting model predictive performance as the small choice sets experiment proved that model performance improves significantly with larger ($N_{train} > 10$) training choice sets. These models and contextual performance evaluation provide a theoretical foundation for using non-parametric and semi-parametric machine learning methods for individual preference modeling.

### 5.2. Future work

A future direction of this research is to apply these preference learning methods in a discrete choice elicitation setting and examine how well these methods can capture a real individual's discrete choice preferences. While the case study captured some ways this work can be applied in a real energy policy scenario, the application of this work can be extended through a decision support system that dynamically learns an individual's discrete choice preferences, as seen in [75]. In addition to tailoring decision recommendations by dynamically learning an individual's preference, the trained semi- and non-parametric models can also be used as a proxy for the individual's choice-making process, in a context where such a process cannot directly be observed otherwise.

Another extension of this work is to compare the interpretability of the non-parametric models against the parametric and semi-parametric models. Common model explainability techniques, such as SHAP and LIME, can be applied to extract model feature importance, thereby offering insights into how each attribute contributes to the individual's choice-making process. A study on whether incorporating model interpretability metrics in the decision support system alters an individual's decision-making paradigm would be valuable.

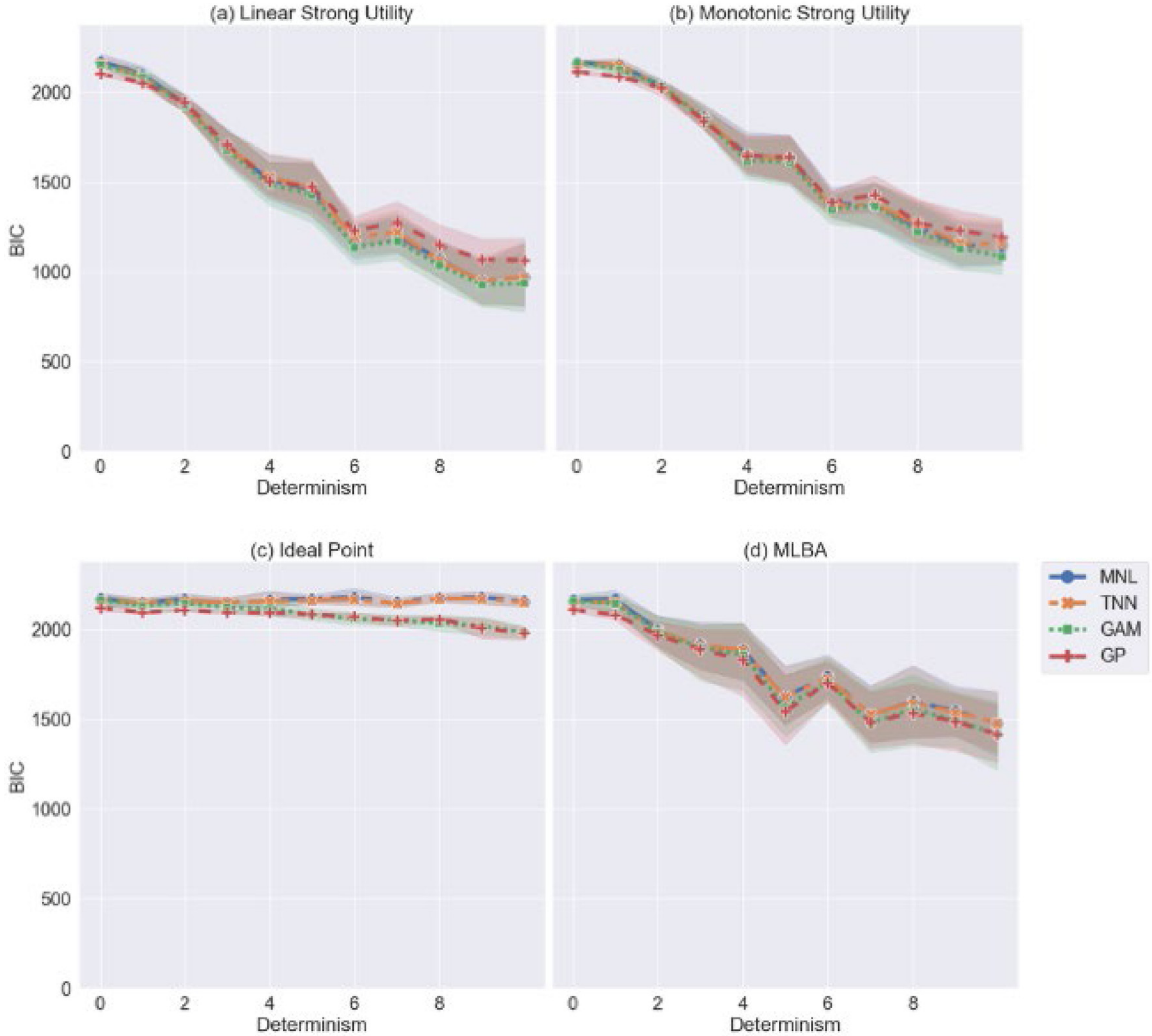

**Fig. 9.** Bayesian information criterion for the determinism experiment.
Note: The scores are averaged across all Monte Carlo trials. The shaded region displays two standard deviations within all brier scores across the Monte Carlo trials for each determinism coefficient. Increasing probabilistic choice rules' determinism decreases model prediction error.

The four models assessed in this study were chosen for their relevance to both choice modeling and machine learning literature, and for the range of their model flexibility. However, the models assessed are not comprehensive of the entirety of machine learning models applicable to discrete choice modeling. Valuable future studies include assessing these model performances against modern ensemble methods, such as random forest or XGBoost.

## 6. Conclusion

Discrete choice modeling applications within the public policy domain typically assume a parametric, often linear, functional form for ease of interpretability [18,76,77]. This study reflects the common challenges behind translating policy alternatives into a discrete choice preference elicitation, including a large set of decision attributes considered, low choice sets availability, and heterogeneous degrees of randomness in an individual's choice preference. This study provides a theoretical foundation for semi- and non-parametric machine learning model selection by demonstrating that these models generally outperform parametric models, especially the popular multinomial logistic regression model, across these challenging contexts. This work can be applied to improve discrete preference recovery among individual policy stakeholders. A future extension of this work includes evaluating these machine learning models on individual preference choice sets elicited from policy stakeholders as a decision support system, assessing the interpretability of the models, and evaluating against ensemble machine learning methods.

## Declaration of competing interest

All authors declares that they have no conflicts of interest.

## Acknowledgments

This research was supported by National Science Foundation's Division of Social and Economic Sciences grant #2049333 and the NSERC PGS-D Scholarship. The authors thank Dr. Baruch Fischhoff at Carnegie Mellon University for his continuous comments, insights, and edits that made this work possible.

## Appendix A. Supplementary data

Supplementary material related to this article can be found online at https://doi.org/10.1016/j.dajour.2025.100668.

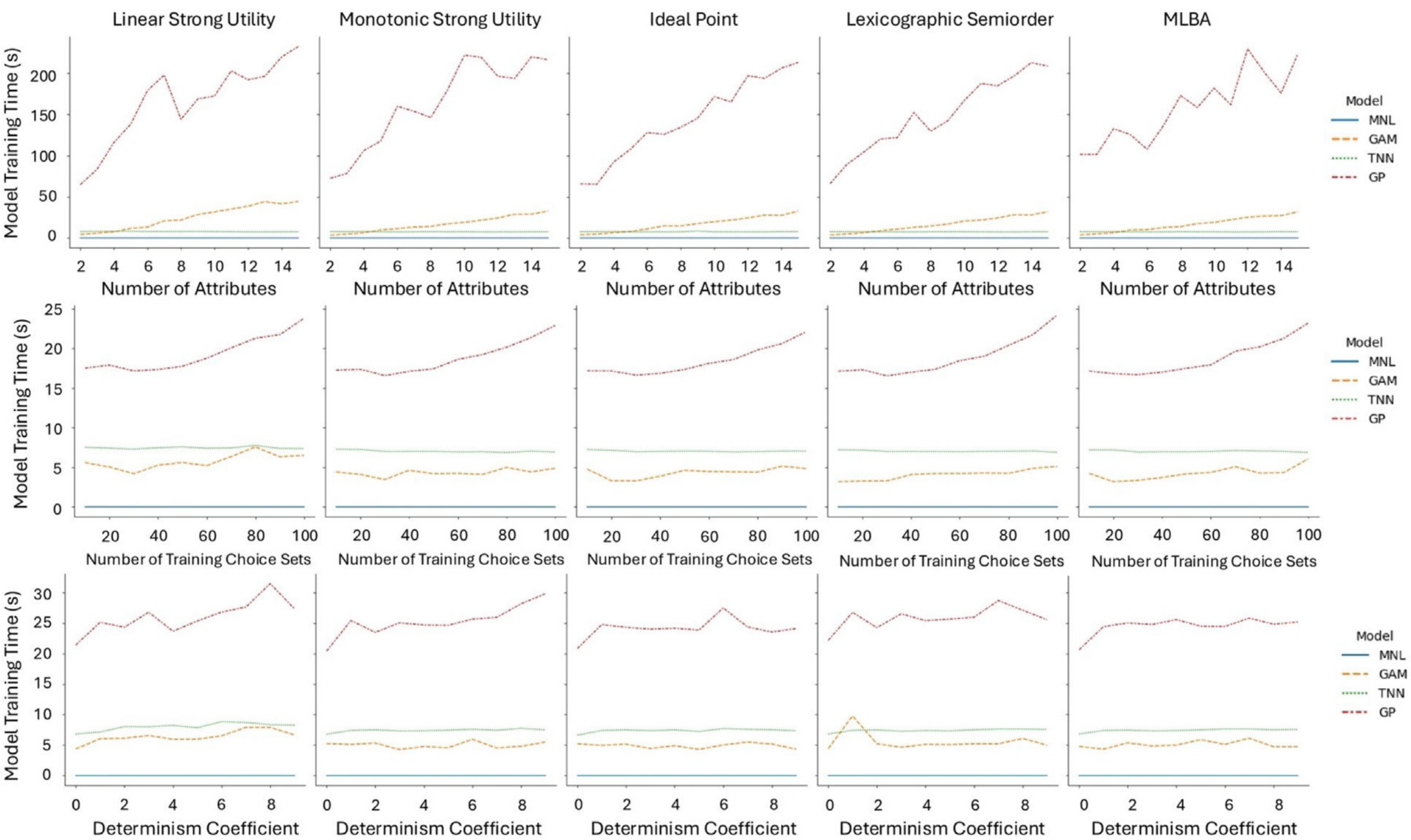

**Fig. 10.** Model training time for each simulation experiment.

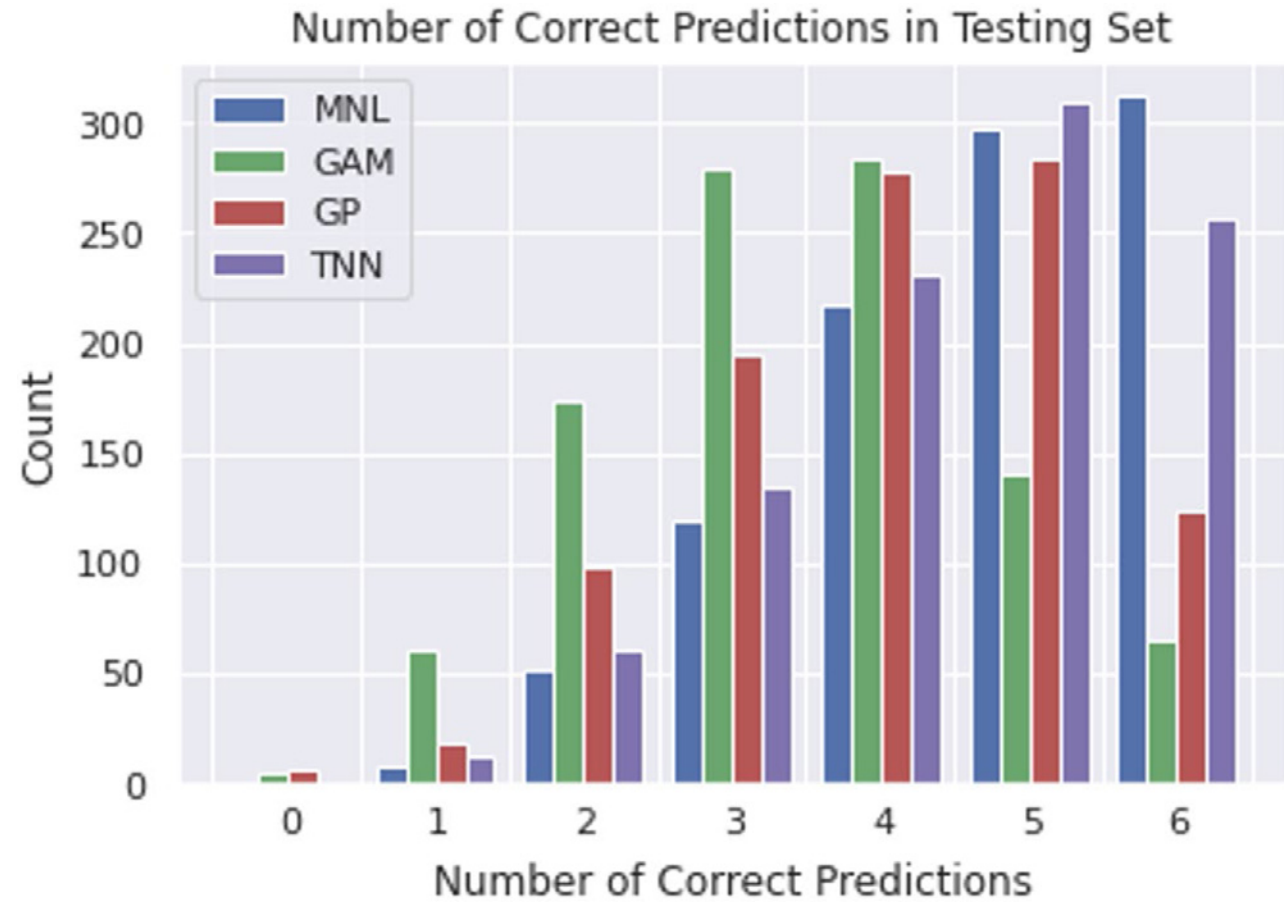

**Fig. 11.** Histogram of correct model predictions for the testing choice sets using the [18] energy policy discrete choice sets.

## Data availability

Data will be made available on request.